\pdfoutput=1
\documentclass[twoside]{article}

\usepackage[accepted]{aistats2024}

\usepackage[round]{natbib}

\usepackage{amsmath,amsfonts,bm}
\usepackage{mathtools}

\newcommand{\eqdef}{\coloneqq}
\newcommand{\range}[2]{\{#1,\dots,#2\}}

\newcommand\thsnd[1]{#1\thinspace000}

\DeclareMathOperator{\ESS}{ESS}
\DeclareMathOperator{\MLP}{MLP}

\newcommand{\x}{\mathbf{x}}

\newcommand{\pd}{p_{\text{data}}}
\newcommand{\ptheta}{p_{\theta}}
\newcommand{\discr}{d_{\phi}}
\newcommand{\xonet}{\x_{\sigma(\leq t)}}
\newcommand{\xt}{x_{\sigma(t)}}
\newcommand{\xtone}{\x_{\sigma(<t)}}
\newcommand{\Ct}{C_t(\xtone)}
\newcommand{\Wt}{W_t(\{\xtone, \xt\})}
\newcommand{\Wtone}{W_{t-1}(\xtone)}
\newcommand{\gammat}{\gamma_t(\xonet)}
\newcommand{\gammatone}{\gamma_{t-1}(\xtone)}

\newcommand{\loss}{\mathcal{L}}

\newcommand{\E}[2]{\mathbb{E}_{#1} \left[ #2 \right]}

\def\eqref#1{equation~\ref{#1}}

\def\1{\bm{1}}

\DeclareMathAlphabet{\mathsfit}{\encodingdefault}{\sfdefault}{m}{sl}
\SetMathAlphabet{\mathsfit}{bold}{\encodingdefault}{\sfdefault}{bx}{n}

\newcommand{\pdata}{p_{\rm{data}}}

\usepackage{hyperref}
\usepackage{cleveref}
\usepackage{url}
\usepackage{algorithm}
\usepackage{algorithmic}
\usepackage{graphicx}
\usepackage{color}
\usepackage{multirow}

\usepackage{xspace}
\usepackage{amssymb}

\newcommand{\suppmat}{appendix\xspace}
\newcommand{\wrt}{w.r.t.}

\usepackage{tikz}
\usetikzlibrary{decorations.pathreplacing,calligraphy,calc,positioning, tikzmark, intersections}
\tikzset{dot/.style = {
    shape  = circle,
    draw   = black,
    fill = black,
    minimum size = 0.2cm
}}

\tikzset{squarenode/.style = {
        shape  = rectangle,
        draw   = black,
        minimum height = 10cm,
        minimum width  = 10cm
}}

\usepackage{pgfplots}

\begin{document}
\twocolumn[ %

\aistatstitle{Discriminator Guidance for Autoregressive Diffusion Models}

\aistatsauthor{Filip Ekström Kelvinius \And Fredrik Lindsten}
\aistatsaddress{
Linköping University \And Linköping University
}
]

\begin{abstract}
We introduce discriminator guidance in the setting of Autoregressive Diffusion Models. The use of a discriminator to guide a diffusion process has previously been used for continuous diffusion models, and in this work we derive ways of using a discriminator together with a pretrained generative model in the discrete case. First, we show that using an optimal discriminator will correct the pretrained model and enable exact sampling from the underlying data distribution. Second, to account for the realistic scenario of using a sub-optimal discriminator, we derive a sequential Monte Carlo algorithm which iteratively takes the predictions from the discriminator into account during the generation process. We test these approaches on the task of generating molecular graphs and show how the discriminator improves the generative performance over using only the pretrained model.
\end{abstract}

\section{INTRODUCTION}
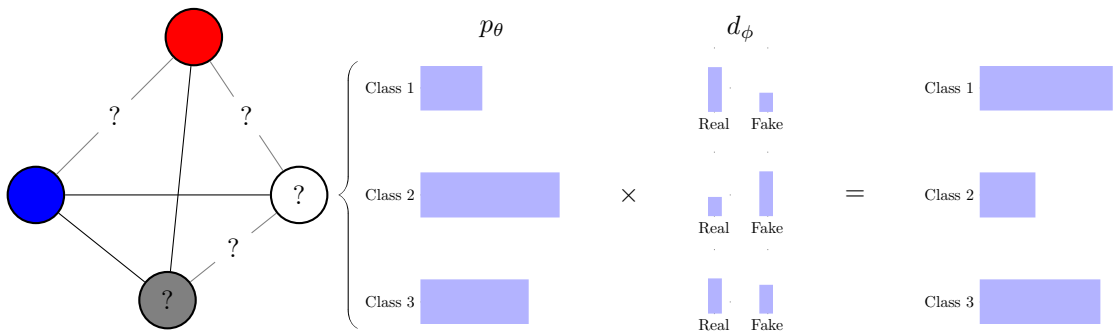
\begin{figure*}
\begin{center}
    \begin{tikzpicture}

        \begin{scope}[scale=0.7, local bounding box=graph]
            \begin{scope}[every node/.style={circle,thick, draw, minimum size=0.75cm}]
                \node[fill=blue] (A) at (0,3) {};
                \node[fill=red] (B) at (3,6) {};
                \node[fill=gray] (C) at (2.5,1) {?};
                \node (D) at (5,3) {?} ;
            \end{scope}

            \begin{scope}[
              every node/.style={fill=white,circle},
              ]
                \path[draw=gray] (A) edge node {?} (B);
                \path (A) edge (C);
                \path (A) edge (D);
                \path (B) edge (C);
                \path[draw=gray] (B) edge node {?} (D);
                \path[draw=gray] (C) edge node {?} (D);
            \end{scope}
        \end{scope}

        \draw [decorate, decoration = {calligraphic brace, amplitude=7pt}] ([shift={(0.5,-1.5)}] D.south east) -- ([shift={(0.5,1.5)}] D.north east);
        
        \begin{scope}[scale=0.6,local bounding box=pthetabar,
        shift={($(D.east) + (0.7,0)$)}, anchor=west, scope anchor]
                \begin{axis}[
            xbar=0.1cm,
            bar width=1.0cm,
            major grid style={draw=white},
            xmin=0.0, xmax=1.0,
            axis x line=none,
            y axis line style={opacity=0},
            tickwidth=0pt,
            symbolic y coords={
               Class 3,Class 2,Class 1},
           ytick=data,
        ]
        \addplot [draw=none, fill=blue!30] coordinates {
          (0.2,Class 1) 
          (0.45,Class 2)
          (0.35,Class 3)};
          \end{axis}
        \end{scope}

        \node[] (ptheta) at ([shift={(-0.7,0.5)}] pthetabar.north) {$p_\theta$};
        
        \begin{scope}[local bounding box=plus, shift={($(pthetabar.east) + (-1.6,0)$)}, anchor=west, scope anchor]
            \node[] () at (0,0) {$\times$};
        \end{scope}

        \begin{scope}[scale=0.6,local bounding box=class2w,
        shift={($(plus.east) + (1.0,0)$)}, anchor=west, scope anchor]
            \begin{axis}[
                ybar=0.1cm,
                height=3cm, width=3cm,
                major grid style={draw=white},
                ymin=0.0, ymax=1.0,
                axis y line=none,
                x axis line style={opacity=0},
                tickwidth=0pt,
                symbolic x coords={
                   Real,Fake},
               xtick=data,
            ]
            \addplot [draw=none, fill=blue!30] coordinates {
            (Real, 0.3)
            (Fake, 0.7)
            };
        \end{axis}
        \end{scope}
        
        \begin{scope}[scale=0.6,local bounding box=class1w,
        shift={($(class2w.north) + (0.0, 0.4)$)}, anchor=south, scope anchor]
            \begin{axis}[
                ybar=0.1cm,
                height=3cm, width=3cm,
                major grid style={draw=white},
                ymin=0.0, ymax=1.0,
                axis y line=none,
                x axis line style={opacity=0},
                tickwidth=0pt,
                symbolic x coords={
                   Real,Fake},
               xtick=data,
            ]
            \addplot [draw=none, fill=blue!30] coordinates {
            (Real, 0.7)
            (Fake, 0.3)
            };
        \end{axis}
        \end{scope}

        \begin{scope}[scale=0.6,local bounding box=class3w,
        shift={($(class2w.south) + (0.0,-0.25)$)}, anchor=north, scope anchor]
            \begin{axis}[
                ybar=0.1cm,
                height=3cm, width=3cm,
                major grid style={draw=white},
                ymin=0.0, ymax=1.0,
                axis y line=none,
                x axis line style={opacity=0},
                tickwidth=0pt,
                symbolic x coords={
                   Real,Fake},
               xtick=data,
            ]
            \addplot [draw=none, fill=blue!30] coordinates {
            (Real, 0.55)
            (Fake, 0.45)
            };
        \end{axis}
        \end{scope}

        \begin{scope}[scale=1.0, anchor=center, scope anchor, shift={($(class1w.center|-ptheta.center) + (0.0,0.0)$)}]
            \node[] at (0,0) {$\discr$};
        \end{scope}
        
        \begin{scope}[scale=1.0,local bounding box=equal,
        shift={($(plus.east) + (2.5, 0.0)$)}, anchor=west, scope anchor]
        \node[] () at (0,0) {$=$};
        \end{scope}

        \begin{scope}[scale=0.6,local bounding box=discrbar,
        shift={($(equal.east) + (1.0,0)$)}, anchor=west, scope anchor]
                \begin{axis}[
            xbar=0.1cm,
            bar width=1.0cm,
            major grid style={draw=white},
            xmin=0.0, xmax=1.0,
            axis x line=none,
            y axis line style={opacity=0},
            tickwidth=0pt,
            symbolic y coords={
               Class 3,Class 2,Class 1},
           ytick=data,
        ]
        \addplot [draw=none, fill=blue!30] coordinates {
          (0.43,Class 1) 
          (0.18,Class 2)
          (0.39,Class 3)};
          \end{axis}
        \end{scope}
        
    \end{tikzpicture}
\end{center}
\caption{\label{fig:graphical_abstract} Illustration of the ARDG method when applied to graphs. The variables $x_t$ (nodes and edges in the graph) are one by one assigned values, where nodes and edges with "?" correspond to variables which have not yet been assigned. Assignment of the white node could be done by sampling from the conditional distribution $\ptheta(x_{\sigma(t)} |\x_{\sigma(<t)})$, which has been learnt by a neural network. In our method, however, a separate discriminator, $\discr$, has been trained to distinguish between real and fake (generated by the generative model $\ptheta$) samples. With the help of this discriminator, we can correct the distribution $\ptheta(x_{\sigma(t)} | \x_{\sigma(<t)})$ so that it becomes closer to the true underlying data distribution $\pdata(x_{\sigma(t)}|\x_{\sigma(<t)})$.}
\end{figure*}

Diffusion models \citep{sohl-dickstein_deep_2015, ho_denoising_2020-1, song_score-based_2021} have in the last couple of years received significant attention, and many improvements and extensions of the original works have been proposed. A recent work by \citet{kim_refining_2023-1} introduced \emph{discriminator guidance}: a way of improving the score estimation of a pretrained score-based diffusion model by training a discriminator, and during generation combine the pretrained score model with the gradient of the discriminator to obtain an improved estimation of the score. Empirically, this approach improves sample quality. 

Score-based models operate on continuous data. One formulation of a diffusion model for \emph{discrete} data is the Autoregressive Diffusion Model (ARDM) by \citet{hoogeboom_autoregressive_2022} which builds on the Order-Agnostic Autoregressive Model (OA-ARM) by \citet{uria_deep_2014}. This is a significantly different formulation of a diffusion model compared to the score-based model, and it is therefore not straightforward to use  the existing formulation of discriminator guidance for this type of model. However, given the improved performance obtained by incorporating discriminator guidance in the continuous setting, it is of interest to develop such techniques 
also for the ARDM. 

To address this challenge, we formulate the following algorithms for incorporating a discriminator into the sampling from an ARDM:
\begin{itemize}
\item Autoregressive Discriminator Guidance (ARDG), which is the most similar to the continuous case by \citet{kim_refining_2023-1}: the conditional distribution predicted by the ARDM is corrected by a discriminator, which we show enables exact sampling under the (in practice unrealistic) assumption that we can find a perfect discriminator.
\item Sequential Monte Carlo (SMC) algorithms which build on discriminator guidance, but aims at mitigating the accumulation of errors that can occur due to imperfect discriminators for the intermediate sampling steps.
This is made possible by the sequential correction of intermediate target distributions through the propagation, weighting, and resampling steps of SMC.
\end{itemize}

Using these different approaches, we empirically verify that discriminator guidance improves the generative sample quality over regular ARDMs in the setting of generating molecular graphs. Our code is available online\footnote{\url{https://github.com/filipekstrm/graph_ardm}}.

\section{BACKGROUND}
Consider the task of generating a $D$-dimensional vector $\x = (x_1, x_2, \dots, x_D)$, where each variable $x_i$ is discrete. To this end, we will use an Autoregressive Diffusion Model (ARDM) \citep{hoogeboom_autoregressive_2022} in combination with a discriminator. This section therefore introduces background on the ARDM, discriminator guidance for \emph{continuous} data, and some additional related work.

We write $\pdata(\x)$ for the true data distribution, and $\ptheta(\x)$ for the generative model, which is effectively an estimate of $\pdata(\x)$ constructed using neural networks with parameters $\theta$.

\subsection{Autoregressive Diffusion Models}
To introduce ARDMs, we start from the standard autoregressive model (ARM). For an ARM, the data log-likelihood $\log \ptheta(\x)$ is factorized as
\begin{align}
    \log \ptheta(\x) = \sum_{t=1}^D \log \ptheta(x_t | \x_{1:t-1}),
\end{align}
where $\x_{1:t-1} \eqdef (x_1, \dots, x_{t-1})$.
Under the autoregressive model, data can be generated via ancestral sampling, i.e., sample the variables $x_t$ conditioned on the previously sampled values $\x_{1:t-1}$.

However, the autoregressive model assumes a certain order of the variables $x_t$. In the \emph{order agnostic} autoregressive model (OA-ARM) \citep{uria_deep_2014}, the order is not viewed as fixed, but a random variable drawn from the set of all permutations $S_D$ of the indices $\{1,\dots, D\}$. Denoting the permutation as $\sigma = (\sigma(1), \dots, \sigma(D))$, the data log-likelihood for OA-ARM, given the order $\sigma$, is written as
\begin{align}
    \log \ptheta(\x|\sigma) = \sum_{t=1}^D \log \ptheta(x_{\sigma(t)} | \x_{\sigma(<t)}). \label{eq:cond-loglik}
\end{align} 
Here, and in what follows, we use the shorthand notation
$\sigma(<t) \eqdef (\sigma(1), \dots, \sigma(t-1))$ for brevity, to denote the first $t-1$ elements of the permutation.
The full data log-likelihood can then be written as the expectation over all orders
\begin{align}
    \log \ptheta(\x) = \log \E{\sigma}{\ptheta(\x |\sigma)}.
\end{align}
To formulate ARDM, \citet{hoogeboom_autoregressive_2022} use the OA-ARM formulation in a diffusion model context: by having a generative process according to \Cref{eq:cond-loglik}, i.e., pick an order randomly and then one by one assigning values to the variables (or "unmasking" them) according to this order, the noising process corresponds to randomly "masking out" variables one by one.

An OA-ARM or ARDM can then be trained by maximizing the log-likelihood as
\begin{align}
 \log {}&\ptheta(\x) = \log \E{\sigma}{\ptheta(\x |\sigma)} \nonumber 
 = \E{\sigma}{\log \ptheta(\x |\sigma)} \nonumber \\
 &= \E{\sigma}{\sum_{t=1}^D \log \ptheta(x_{\sigma(t)} | \x_{\sigma(<t)})} \nonumber \\
 &= \E{\sigma}{D \E{t}{\log \ptheta(x_{\sigma(t)} | \x_{\sigma(<t)})}} \nonumber \\
 &= D\E{t}{\E{\sigma(<t)}{\E{\sigma(t)|\sigma(<t)}{\log \ptheta(x_{\sigma(t)} | \x_{\sigma(<t)})}}}.
 \label{eq:ardm-factorization}
\end{align}
It can be noted that, contrary to \citet{uria_deep_2014} and \citet{hoogeboom_autoregressive_2022}, we have not relied on Jensen's inequality for moving the $\log$ inside the expectation (first equality). The reason for this is that a different $\sigma$ merely leads to a different factorization of the same likelihood, and hence, $\log \ptheta(\x |\sigma) = \log \ptheta(\x), \ \forall \sigma$.\footnote{Put differently, the true likelihood is independent of $\sigma$ and the introduction of $\sigma$ as a latent variable only affects the training of the model through the stochastic approximation of the expectation in \Cref{eq:ardm-factorization}.}

As $\sigma$ is drawn uniformly from the set of all permutations, the inner-most expectation can be rewritten as a sum over all masked elements, and training an ARDM is done by maximizing the log-likelihood
\begin{align}
    &\log \ptheta(\x) = \nonumber \\
    &
    D\E{t}{\E{\sigma(<t)}{\frac{1}{D-t+1}\sum_{k\notin \sigma(<t)}\log \ptheta(x_{k} | \x_{\sigma(<t)})}}.
\end{align}

\subsection{Discriminator Guidance for Score-based Diffusion Models}
In score-based diffusion models \citep{song_generative_2019, song_score-based_2021}, a neural network $\textbf{s}_\theta$ is trained to estimate the score $\nabla\log \pdata(\x)$, where $\x$ is now a continuous variable. The trained neural network $\textbf{s}_{\theta}$ (implicitly) defines a distribution $\ptheta$. If the learnt score-function $\textbf{s}_\theta$ deviates from the real score $\nabla \log \pdata$, there will be a gap between $\pd$ and $\ptheta$. By observing that
\begin{align}
    \nabla \log\pd(\x) &= \textbf{s}_\theta(\x) + \nabla\log\frac{\pd(\x)}{\ptheta(\x)},
\end{align}
\citet{kim_refining_2023-1} introduce a correction term $\nabla\log\frac{\pd(\x)}{\ptheta(\x)}$. This is not tractable, but they explain how it can be estimated by training a discriminator $d_\phi$ to discriminate between real and generated samples: if $d_{\phi}(\x)$ can accurately predict the probability that a sample $\x$ is drawn from the data distribution, then the ratio $\frac{d_\phi (\x)}{1-d_\phi (\x)}$ approximates $\frac{\pd(\x)}{\ptheta(\x)}$ and hence
\begin{align}
    \textbf{c}_\phi(\x) \eqdef \nabla \log \frac{d_\phi (\x)}{1-d_\phi (\x)}
    \approx 
    \nabla\log\frac{\pd(\x)}{\ptheta(\x)}.
\end{align}
This correction term is used together with $\textbf{s}_\theta$ so that the estimated score is $\textbf{s}_\theta(\x) + \textbf{c}_\phi(\x)$, which with a perfectly trained discriminator will correspond to the exact score $\nabla \log \pd(\x)$. 

\subsection{Other Related Work}
\textbf{Discrete Diffusion Models:}
Some early works on discrete diffusion models are Multinomial Diffusion \citep{hoogeboom_argmax_2021} and D3PM
\citep{austin_structured_2021}. In both these cases, the noise process consists of the variables independently transitioning to different states. In D3PM, an additional noising process is introduced, absorbing state noise: at each time step, variables are independently and with a certain probability "decayed" into an "absorbed" state. ARDM is a continuous-time generalization of this process: in continuous time, at most one variable at a time will decay into the absorbed state, and the noise process hence boils down to finding the order in which the variables decay. If the variables have the same probability of decaying, the order in which they decay will be a draw from the uniform distribution over all permutations of the indices $\{1,\dots,D\}$. The reverse process is to "unmask" the absorbed variables in the reverse order in which they decayed. This reverse order will also be from the uniform distribution, like in OA-ARM. 

In our work, we focus on models which operate on discrete state spaces. There are, however, formulations of diffusion models for discrete data which use \emph{continuous} diffusion. Examples of these are EDM \citep{hoogeboom_equivariant_2022}, CDCD \citep{dieleman_continuous_2022}, and Bit Diffusion \citep{chen_analog_2023}.

\textbf{Diffusion Models for Graphs:} In this work, we will apply our model to the task of generating graphs. A previous work by \citet{niu_permutation_2020} use continuous diffusion to generate adjacency matrices, and \citet{jo_score-based_2022} use continuous diffusion to generate graphs with node and edge attributes. DiGress \citep{vignac_digress_2023} is an extension of D3PM to graphs, which generates graphs with node and edge attributes and is the first diffusion model for graphs that use discrete noise. 

\textbf{Generative Adversarial Networks:}
Generative adversarial networks (GANs) \citep{goodfellow_generative_2014} are perhaps the class of generative models which is mostly associated with the term "discriminator" as training of GANs is done by training a generative model and discriminator simultaneously in an adversarial manner. In a standard GAN, the discriminator is merely a training artifact and not used for generation. However, follow-up work have found ways of using the discriminator also in the generative process by rejection sampling \citep{azadi_discriminator_2018}, Metropolis--Hasting sampling \citep{turner_metropolis-hastings_2019-1}, and Langevin sampling \citep{che_your_2020}.

\textbf{SMC and Diffusion Models:} Concurrent to our work, \citet{
wu_practical_2023, 
cardoso_monte_2023-1} 
have proposed SMC-based methods for solving inverse problems with diffusion-based priors. The resulting algorithms show some resemblance to our SMC-based samplers, but differ both in that they consider continuous rather than discrete state spaces, and in that they focus on inverse problems and not discriminator guidance.

\section{DISCRIMINATOR GUIDANCE FOR ARDMS}
To formulate a discrete discriminator guidance procedure, we assume we have a pre-trained generative model, $\ptheta$, from which we can sample in an order-agnostic autoregressive manner, i.e.,
\begin{align}
    x_{\sigma(t)} \sim \ptheta(x_{\sigma(t)} | \x_{\sigma(<t)}).
\end{align}
We denote the real data distribution as $\pd$, and train a discriminator $\discr$ by maximizing
\begin{align}
    \loss_{\phi} =
     \mathbb{E}_\sigma  \Bigl[ \mathbb{E}_t \Bigl[ &\E{\pd(\x_{\sigma(\leq t)})}{\log \discr(\x_{\sigma(\leq t)})} + \nonumber \\
    &\E{\ptheta(\x_{\sigma(\leq t)})}{1-\log \discr(\x_{\sigma(\leq t)})}\Bigr]\Bigr]
    .\label{eq:discr_loss}
\end{align}
Note that the discriminator is assumed to accept any 
partial sample $\x_{\sigma(\leq t)}$, $t \in \range{1}{D}$ as input, and return the probability that this is a partially masked sample from the data distribution.

For a fixed generative model $\ptheta$, training the discriminator $\discr$ by minimizing \Cref{eq:discr_loss} has the optimum $\discr^* (\x_{\sigma(\leq t)}) = \frac{\pd(\x_{\sigma(\leq t)})}{\pd(\x_{\sigma(\leq t)}) + \ptheta(\x_{\sigma(\leq t)})}$ \citep{goodfellow_generative_2014}. If we then define
\begin{align}
    W_t(\x_{\sigma(\leq t)}) \eqdef \frac{\discr(\x_{\sigma(\leq t)})}{1-\discr(\x_{\sigma(\leq t)})}, \label{eq:wt-def}
\end{align}
simple algebra gives, for an optimal discriminator $\discr^*$,
\begin{align}
    W_t^*(\x_{\sigma(\leq t)}) =  \frac{\discr^*(\x_{\sigma(\leq t)})}{1-\discr^*(\x_{\sigma(\leq t)})} = \frac{\pd(\x_{\sigma(\leq t)})}{\ptheta(\x_{\sigma(\leq t)})}. \label{eq:wt}
\end{align}
Note that, if we parameterize the discriminator as $\discr(\x_{\sigma(\leq t)}) = \text{sigmoid}(f_{\phi}(\x_{\sigma(\leq t)})) = \frac{\exp(f_\phi(\x_{\sigma(\leq t)}))}{1 + \exp(f_\phi(\x_{\sigma(\leq t)}))}$, it follows that
\begin{align}
    W_t(\x_{\sigma(\leq t)}) 
    = \exp(f_\phi(\x_{\sigma(\leq t)})). \label{eq:W_logit_space}
\end{align}
In the following sections, we outline our sampling algorithms which combine the pretrained model $\ptheta$ and the discriminator $\discr$.

\begin{algorithm}[tb]
   \caption{Autoregressive Discriminator Guidance} %
   \label{alg:discriminator_guidance}
\begin{algorithmic}
   \STATE 
   Sample $D\sim p(D)$
   \COMMENT{For samples with varying number of elements (see \Cref{sec:varying-n})}
   \STATE Sample $\sigma \sim p(\sigma)$
   \STATE Initialize $\x_0 = \emptyset$ \COMMENT{Completely masked}
   \FOR{$t$ in $1:D$}
   \STATE $C_t = 0$ \COMMENT{Normalization constant}
   \FOR{each value of $x_{\sigma(t)}$}
   \STATE Compute $\discr(\{\x_{\sigma(<t)}, x_{\sigma(t)}\})$
   \STATE Compute $W_t(\{\x_{\sigma(<t)}, x_{\sigma(t)}\})$ (\Cref{eq:Wt-def})
   \STATE $C_t \mathrel{+}= W_t(\{\x_{\sigma(<t)}, x_{\sigma(t)}\}) \ptheta(x_{\sigma(t)} | \x_{\sigma(<t)})$
   \ENDFOR
   \STATE Sample \[\xt \sim %
   C_t^{-1}
   W_t(\{\x_{\sigma(<t)}, x_{\sigma(t)}\}) \ptheta(x_{\sigma(t)} | \x_{\sigma(<t)})\]
   \ENDFOR
\end{algorithmic}
\end{algorithm}

\subsection{Autoregressive Discriminator Guidance}
\label{subsec:discr}
We first derive a discriminator guidance algorithm 
which is a discrete counterpart of the continuous case
(i.e., \citet{kim_refining_2023-1}). To this end, we start by rewriting
\begin{align}
\pd(\x_{\sigma(\leq t)}) = \pd(x_{\sigma(t)} | \x_{\sigma(<t)}) \pd(\x_{\sigma(<t)}).
\end{align}
This can be done similarly for $\ptheta(\x_{\sigma(\leq t)})$, and with this, we get
\begin{align}
    W_t^*(\x_{\sigma(\leq t)}) &= \frac{\pd(x_{\sigma(t)} | \x_{\sigma(<t)}) \pd(\x_{\sigma(<t)})}{\ptheta(x_{\sigma(t)} | \x_{\sigma(<t)}) \ptheta(\x_{\sigma(<t)})} \nonumber \\
    &= \frac{\pd(x_{\sigma(t)} | \x_{\sigma(<t)})}{\ptheta(x_{\sigma(t)} | \x_{\sigma(<t)})}W^*_{t-1}(\x_{\sigma(<t)}) \nonumber \\
    &\iff \nonumber 
    \\
    \pd(x_{\sigma(t)} | &\x_{\sigma(<t)}) = \frac{W_{t}^*(\x_{\sigma(\leq t)})}{W_{t-1}^*(\x_{\sigma(<t)})}
    \ptheta(x_{\sigma(t)} | \x_{\sigma(<t)}). \label{eq:pdata_cond}
\end{align}
As $x_{\sigma(t)}$ is discrete, the distribution on the right hand side of \Cref{eq:pdata_cond} can be computed as: for each value of $x_{\sigma(t)}$, evaluate the discriminator $\discr^*(\{\x_{\sigma(<t)}, x_{\sigma(t)}\})$, compute $W_t^*(\{\x_{\sigma(<t)}, x_{\sigma(t)}\})$, and multiply this with $\ptheta(x_{\sigma(t)} |\x_{\sigma(<t)})$. The denominator $W_{t-1}^*(\x_{\sigma(<t)})$ does not depend on $x_{\sigma(t)}$, and can therefore be implicitly computed by normalizing the probabilities $W_{t}^*(\x_{\sigma(\leq t)})\ptheta(x_{\sigma(t)} | \x_{\sigma(<t)})$. Hence, with a perfectly trained discriminator, we can now at each time step sample from the data distribution, $\pd(x_{\sigma(t)} | \x_{\sigma(<t)}) \propto W^*_t(\{\x_{\sigma(<t)}, 
x_{\sigma(t)}\})\ptheta(x_{\sigma(t)} | \x_{\sigma(<t)})$. In other words, we use the discriminator to correct the imperfect predictions of the intermediate conditionals made by $\ptheta$. In practice, we will not have access to a perfect discriminator, but we can use the approximation 
\begin{align}
\label{eq:Wt-def}
W_t(\{\x_{\sigma(<t)}, x_{\sigma(t)}\}) = \frac{\discr(\{\x_{\sigma(<t)}, x_{\sigma(t)}\})}{1-\discr\{\x_{\sigma(<t)}, x_{\sigma(t)}\})},
\end{align}
to obtain a guided sampling process to approximately sample from $\pd$.
We refer to this procedure as Autoregressive Discriminator Guidance (ARDG).
The full procedure can be found in \Cref{alg:discriminator_guidance}, and is illustrated in \Cref{fig:graphical_abstract}.
In the algorithm we have assumed that the generation order (permutation) can be sampled before the main loop: $\sigma\sim p(\sigma)$ (which is typically the case for ARDMs), but it is also possible to sample this one variable at a time $\sigma(t) \sim p(\sigma(t) | \sigma(<t))$ as part of the loop.

\subsection{Sequential Monte Carlo}
\begin{algorithm}[tb]
   \caption{
   Bootstrap particle filter discriminator guidance (BSDG). All operations for $i=\range{1}{N}$
   }
   \label{alg:ourbspf}
\begin{algorithmic}
\STATE Sample $D\sim p(D)$
\COMMENT{For samples with varying number of elements (see \Cref{sec:varying-n})}
\STATE Sample $\sigma \sim p(\sigma)$
\STATE Set $\x_0^i = \emptyset$ and $w_0^i = 1/N$ \COMMENT{Completely masked}
\FOR{$t$ in $1:D$}
\IF{ESS too low}
\STATE Resample $\{\x_{\sigma(<t)}^i\}_{i=1}^N$ and set $w_{t-1}^i \equiv 1/N$
\ENDIF
\STATE Sample $x_{\sigma(t)}^i \sim \ptheta(x_{\sigma(t)} | \x_{\sigma(<t)}^{i})$
and set $\x_{\sigma(\leq t)}^i = (\x_{\sigma(<t)}^{i}, x_{\sigma(t)}^i)$%
\STATE Compute 
$\tilde{w}_t^i = w_{t-1}^i W_t(\x_{\sigma(\leq t)}^i)/W_{t-1}(\x_{\sigma(<t)}^{i})$
\STATE Normalize $w_t^i = \tilde{w}_t^i / \sum_{j=1}^N \tilde{w}_t^j$
\ENDFOR
\STATE Sample $k \sim \text{Categorical}(N, \{w_D^i\}_{i=1}^N)$
\RETURN $\x = \x_{\sigma}^k$
\end{algorithmic}
\end{algorithm}

\begin{algorithm}[tb]
   \caption{
   Fully adapted SMC discriminator guidance (FADG). All operations for $i=\range{1}{N}$
   }
   \label{alg:ourfapf}
   \begin{algorithmic}
\STATE Sample $D\sim p(D)$
\COMMENT{For samples with varying number of elements (see \Cref{sec:varying-n})}
\STATE Sample $\sigma \sim p(\sigma)$
\STATE Set $\x_0^i = \emptyset$ and $w_0^i = 1/N$ \COMMENT{Completely masked}
\FOR{$t$ in $1:D$}
\STATE Compute
\(
C_{t-1}^i = \sum_{x_{\sigma(t)}} \frac{ W_{t}( \{\x_{\sigma(<t)}^i, x_{\sigma(t)}\}) }{ W_{t-1}(\x_{\sigma(<t)}^i) }
\)
\STATE Compute 
$\tilde{w}_t^i = w_{t-1}^i C_{t-1}^i$
\STATE Normalize $w_t^i = \tilde{w}_t^i / \sum_{j=1}^N \tilde{w}_t^j$
\IF{ESS too low}
\STATE Resample $\{\x_{\sigma(<t)}^i, C_{t-1}^i\}_{i=1}^N$ and set $w_{t}^i \equiv 1/N$
\ENDIF
\STATE Sample 
\[
x_{\sigma(t)}^i \sim \frac{1}{C_{t-1}^i}\frac{W_{t}( \{\x_{\sigma(<t)}^i, x_{\sigma(t)}\})}{W_{t-1}(\x_{\sigma(<t)}^i)} \ptheta(x_{\sigma(t)} | \x_{\sigma(<t)}^{i})
\]
and set $\x_{\sigma(\leq t)}^i = (\x_{\sigma(<t)}^{i}, x_{\sigma(t)}^i)$%
\ENDFOR
\STATE Sample $k \sim \text{Categorical}(N, \{w_D^i\}_{i=1}^N)$
\RETURN $\x = \x_\sigma^k$
\end{algorithmic}
\end{algorithm}

A possible drawback with ARDG is that (as any "plain" autoregressive model) it is unable to correct, at some later iteration, for mistakes made at earlier iterations during the generation.
However, intuitively we might expect that the discriminator becomes more accurate for large $t$, since it then has access to more unmasked elements and thus more information.
We therefore propose an extension to ARDG based on Sequential Monte Carlo (SMC; see, e.g., \cite{NaessethLS:2019a}).

The motivation is that SMC can potentially be more resilient to an accumulation of errors. The reason is that SMC has a built-in correction mechanism, in the sense that "errors" in the intermediate target distributions of the algorithm (details below) are corrected for when transitioning from one iteration to the next. Therefore, it is sufficient that the final target distribution is error-free to obtain an algorithm that is consistent, i.e., asymptotically unbiased as the number of SMC samples ("particles") increases \citep{NaessethLS:2019a}. In our setting, this means that as long as the discriminator $\discr(\x)$ which has access to a complete, unmasked sample is (a close approximation to) the optimal classifier, the guided sampling process will asymptotically generate samples from (a close approximation of) the data distribution.

In practice we view the number of SMC particles $N$ as a tuning parameter that can be used to trade computational cost for improved accuracy. Importantly, even in the extreme case $N=1$, the two versions of SMC that we propose below will reduce to standard ARDM and ARDG, respectively. We therefore argue that even a small $N$ can be used to improve the performance compared to the respective baselines.

In the derivation below we condition on a fixed generation order $\sigma$, but do not include this in the notation for brevity.
SMC is a class of methods for sampling from a sequence of "target distributions" $\{\pi_t(\x_{\sigma(\leq t)})\}_{t=1}^D$.
This is done by sequentially generating a collection of $N$ interacting samples (or particles) that can be seen as approximate draws from the targets. The target distributions are assumed to be known only up to normalization, such that
\begin{align*}
    \pi_t(\x_{\sigma(\leq t)}) = \frac{\gamma_t(\x_{\sigma(\leq t)})}{Z_t}
\end{align*}
where $\gamma_t(\x_{\sigma(\leq t)})$ can be evaluated but $Z_t$ may be intractable. In many cases---including the application studied in this paper---the actual distribution of interest corresponds to the final target distribution $\pi_D(\x_{\sigma(\leq D)})$
and the \emph{intermediate} target distributions, $\{\pi_t\}_{t=1}^{D-1}$ can be viewed as auxiliary quantities. Importantly, as mentioned above, the algorithm is consistent under weak assumptions regardless of the choice of these intermediate targets, which can thus be seen as design variables.

In what follows we will derive two versions of SMC-based discriminator guidance. Both are based on the same sequence of unnormalized target distributions,
\begin{align}
    \gamma_t(\x_{\sigma(\leq t)}) = W_t(\x_{\sigma(\leq t)}) \ptheta(\x_{\sigma(\leq t)})
    \label{eq:gamma-def}
\end{align}
for $t=\range{1}{D}$, where $W_t$ is defined in  
\Cref{eq:wt-def}.
This means that, at iterations $t=D$, samples are approximately from $\pdata(\x)$ according to \Cref{eq:wt}.

From the definition of $\gamma_t$ in \Cref{eq:gamma-def} it follows that we can write
\begin{align}
    \gamma_t(\x_{\sigma(\leq t)}) = \frac{W_t(\x_{\sigma(\leq t)})
    \ptheta(x_{\sigma(t)} | \x_{\sigma(<t)})
    }{W_{t-1}(\x_{\sigma(<t)})}  \gamma_{t-1}(\x_{\sigma(<t)}),
\end{align}
which can be interpreted as follows:
\begin{enumerate}
    \item The conditional distribution $\ptheta(x_{\sigma(t)} | \x_{\sigma(<t)})$ takes the role of a prior transition from iteration $t-1$ to $t$,
    \item The ratio $W_t(\x_{\sigma(\leq t)}) / W_{t-1}(\x_{\sigma(<t)})$ takes the role of a likelihood term correcting for the discrepancies between the target at iteration $t-1$ and $t$. 
\end{enumerate}
Using this interpretation we can readily obtain the expressions for two commonly used versions of SMC adapted to our setting: the bootstrap SMC and the 
fully adapted SMC, respectively. Full derivations are given in the \suppmat.

\subsubsection{Bootstrap Discriminator Guidance} %
We initialize a set of $N$ particles $\x_0^i = \emptyset$ (completely masked) and corresponding  \emph{importance weights} $w_0^i = 1/N$.
For a bootstrap SMC we use the prior transition, i.e., our pretrained generative model, as a proposal distribution to propagate samples from one iteration to the next. That is, at iteration $t$ we sample
\begin{align}
    \label{eq:bspf-transition}
    x_{\sigma(t)}^i &\sim \ptheta (x_{\sigma(t)} | \x_{\sigma(<t)}^i), & i&=\range{1}{N}.
\end{align}
The samples are extended as $\x_{\sigma(\leq t)}^i = (\x_{\sigma(<t)}^i, x_{\sigma(t)}^i)$.

The importance weights are then updated using the "likelihood", i.e.,
\begin{align}
    w_t^i \propto \frac{W_t(\x_{\sigma(\leq t)}^i)}{W_{t-1}(\x_{\sigma(<t)}^i)} w_{t-1}^i,
\end{align}
and are normalized to sum to one (hence the proportionality sign).

A key concept in SMC is \emph{resampling}, which allows the algorithm to focus on the more promising (high weight) samples and discard the less promising (low weight) ones. This is done by monitoring the Effective Sample Size (ESS), defined as
\begin{align}
    \ESS \eqdef \frac{1}{\sum_{i=1}^N (w_t^i)^2}.
\end{align}
If the ESS becomes too low (say, $\ESS < N/2$), then we generate a new set of particles by sampling with replacement from the current set of particles, with probabilities given by the importance weights. After resampling, the importance weights are reset to be $1/N$.

The Bootstrap Discriminator Guidance (BSDG) procedure for discriminator guidance is summarized in \Cref{alg:ourbspf}. Note that, in the extreme case $N=1$, the procedure reduces to simply applying the pre-trained ARDM without any guidance. $N$ can thus be seen as a tuning-parameter that can be used to improve the quality of generated samples at the cost of increased generation order, and even for $N\gtrsim 1$ we obtain a functional method.

\subsubsection{Fully Adapted Discriminator Guidance}
An alternative to bootstrap SMC is to "adapt" the transition probability for extending the particles as well as the resampling probabilities to the current target distribution. If this is done in a locally optimal way, the method is referred to as fully adapted SMC \citep{pitt_filtering_1999, NaessethLS:2019a}.

Similarly to how \Cref{eq:bspf-transition} is viewed as sampling from the prior, the locally optimal transition corresponds to sampling from a posterior. With our interpretation of the weight ratio as a likelihood, this means that we sample $x_{\sigma(t)}^i$ from a distribution proportional to
\begin{align}
    \label{eq:fapf-transition}
    \frac{W_{t}( \{\x_{\sigma(<t)}^i, x_{\sigma(t)}\})}{W_{t-1}(\x_{\sigma(<t)}^i)}
    \ptheta(x_{\sigma(t)} | \x_{\sigma(<t)}^i),
\end{align}
i.e., the product of the likelihood and the prior.
Note that this is the same transition probability as used by ARDG; cf \Cref{eq:pdata_cond}.

Using this transition kernel, the importance weights are updated as
\begin{align}
    \label{eq:fapf-weights}
    w_{t}^i \propto w_{t-1}^i \times
    \sum_{x_{\sigma(t)}} \frac{W_{t}( \{\x_{\sigma(<t)}^i, x_{\sigma(t)}\})}{W_{t-1}(\x_{\sigma(<t)}^i)}
    \ptheta(x_{\sigma(t)} | \x_{\sigma(<t)}^i),
\end{align}
where the sum is the normalizing constant ("marginal likelihood") of  \Cref{eq:fapf-transition}.
Similarly to above, the weights are normalized to sum to one.

Considering \Cref{eq:fapf-weights} it can be noted that the importance weights are independent of the realized values $\{x_{\sigma(t)}^i\}$ and only depend on particles up to iteration $t-1$. This is a consequence of the perfect adaptation of the transition kernel to the current target $\gamma_t$. However, this also opens up for a one-step look-ahead in the resampling step of the fully adapted SMC. That is, we resample the particles from iteration $t-1$ using the weights in \Cref{eq:fapf-weights} \emph{before} extending the particles to iteration $t$ according to \Cref{eq:fapf-transition}.
When the resampling is performed in this way, we obtain new importance weights $w_t^i = 1/N$; if we do not resample at some iteration we instead compute weights according to \Cref{eq:fapf-weights}.

The resulting algorithm for Fully Adapted Discriminator Guidance (FADG) can be found in \Cref{alg:ourfapf}. In the extreme case $N=1$ the method reduces to the basic ARDG procedure (\Cref{alg:discriminator_guidance}). Hence, similarly to how BSDG can be seen as a generalization of (unguided) ARDM, the FADG can be seen as a generalization of ARDG.

\subsection{Generating Samples of Varying Dimension}\label{sec:varying-n}
We will use ARDM with discriminator guidance for generating graphs. These can have different number of nodes, $n$, 
which in turn means that the dimension $D$ of generated samples is random.
Hence, we are interested in generating from the joint distribution $p(\x, D)$. Similarly to DiGress \citep{vignac_digress_2023}, we enable this by using the factorization $p_\theta(\x, D) = p(D)p_\theta(\x|D)$, where we set $p(D)$ to the empirical distribution computed from the training set.
Specifically, in the case of graphs we sample first the number of nodes $n$ from its empirical distribution, and then, for the case of undirected graphs without self-loops,  compute $D=n + n(n-1)/2$, corresponding to the total number of nodes and edges (see \Cref{sec:experiments} for details). Formally we condition on $D$ in the generative procedures presented above, i.e., we target the distribution $p(\x |D)$ instead of $p(\x)$, but we omit this conditioning for notational simplicity.

Furthermore, in the SMC algorithms we condition on the permutation $\sigma$, so that the generation order is shared between particles. That is, we first simulate $D \sim p(D)$ and $\sigma \sim p(\sigma | D)$ and apply SMC to target $\ptheta(\x | D, \sigma)$. To generate a single sample we then randomly select one of the $N$ particles with probabilities given by their importance weights.\footnote{Alternatively, depending on the application, we can keep all $N$ samples and average over them when computing expectations \wrt~ the generative distribution, resulting in a reduction of Monte Carlo variance.}
To generate multiple samples we repeat this entire procedure, sampling a new pair $(D, \sigma)$ for each round.

\subsection{Computational Complexity}
Adding discriminator guidance will inevitably add an additional computational cost. ARDM requires $D$ evaluations of the generative model. ARDG requires evaluating the discriminator for each possible value of $\xt$, involving $(1+d)D$ network evaluations, with $d$ the dimension of $\xt$ (i.e., for each $t$, one evaluation of the generative model, and then $d$ evaluations of the discriminator). FADG, which samples from the same proposal as ARDG, will add to the computational complexity by a factor of the number of particles, $N$, i.e., it requires $N(1+d)D$ forward passes through the networks.

BSDG on the other hand use the standard ARDM as proposal, and just a single forward pass through the discriminator per sample step $t$ for weighting the particles. This means that it requires $2ND$ evaluations of the networks. For high dimensional $\xt$, the number of network evaluations in BSDG can thus be lower than that of standard ARDG if $N < \frac{(1+d)}{2}$.

\begin{table}[t]
\caption{\label{tab:qm9-results}Evaluation metrics on the QM9 dataset, when using different generation orders and types of Discriminator Guidance (DG). ARDM is a standard autoregressive diffusion model (no guidance) and ARDG, BSDG, and FADG are the three methods proposed in this paper. ARDM* is a standard ARDM that has been trained for twice as long to match the extra training time required for training the discriminator.}
\begin{center}
\begin{tabular}{clcccc}
                        &           
                        & Val. %
                        & Uniq. %
                        & Atm.S %
                        & Mol.S %
                        \\
                   & Model      & (\%)$\uparrow$& (\%)$\uparrow$ & (\%)$\uparrow$ & (\%)$\uparrow$    \\ \hline\hline
                        & Test data    & 97.9          & 100            & 98.6           & 87.4              \\ 
                        & DiGress    & 95.4          & 97.6           & 98.1           & 79.8              \\
                        \hline
\multirow{5}{*}{\rotatebox[origin=c]{90}{Uniform}} 
    & ARDM                 & 88.0     & 99.9       & 96.1       & 55.7       \\
    & ARDM*           & 89.7     & 99.6         & 96.0         & 55.6   \\
    & ARDG            & 88.3     & 99.7       & 96.7       & 66.3       \\
    & BSDG               & 97.4     & 99.4       & 98.6       & 87.8       \\
    & FADG               & 96.7     & 99.5       & 98.7       & 88.1       \\
    \hline
\multirow{5}{*}{\rotatebox[origin=c]{90}{NEsN}}    
    & ARDM                  & 95.4     & 99.9       & 97.3       & 74.2       \\
    & ARDM*            & 96.8     & 99.9       & 97.6       & 78.5      \\
    & ARDG            & 95.3     & 99.9       & 97.4       & 76.9       \\
    & BSDG               & 98.2     & 99.9       & 98.6       & 88.4       \\
    & FADG               & 98.1     & 99.5       & 98.6       & 87.8       \\
    \hline
\multirow{5}{*}{\rotatebox[origin=c]{90}{NsEs}}    
    & ARDM                  & 92.5     & 99.6       & 97.4       & 77.3       \\
    & ARDM*           & 95.4     & 99.7       & 98.0       & 82.4       \\
    & ARDG            & 95.0     & 99.9       & 97.7       & 81.0       \\
    & BSDG               & 97.4     & 99.7       & 98.7       & 88.6       \\
    & FADG               & 97.2     & 99.4       & 98.6       & 88.0       \\
    \hline
\end{tabular}
\end{center}
\end{table}

\begin{table*}[t]
\caption{\label{tab:moses-results}Evaluation metrics on the MOSES dataset, when using different types of Discriminator Guidance (DG). For results for different generation orders, see the \suppmat.}
\begin{center}
\begin{tabular}{clcccccccc}
        &                    & Validity      & Uniqueness     & Novelty        & Filters & FCD         & SNN         & Frag        & Scaf           \\
Order   & DG                 &(\%)$\uparrow$ & (\%)$\uparrow$ & (\%)$\uparrow$ & (\%)$\uparrow$ & $\downarrow$ & $\uparrow$ & $\uparrow$ & $\uparrow$  \\ 
\hline\hline
\multirow{5}{*}{\rotatebox[origin=c]{90}{Uniform}} 
        & ARDM                  & 82.2          & 100            & 97.2           & 94.9   & 2.937       & 0.483       & 0.993       & 0.067          \\
        & ARDM*           & 82.6          & 100            & 97.0           & 95.6   & 3.153         & 0.494       & 0.991       & 0.050          \\
        & ARDG            & 80.5          & 100            & 95.0           & 95.3   & 2.705       & 0.502       & 0.994       & 0.127          \\
        & BSDG               & 85.9          & 100            & 92.4           & 98.4   & 2.609       & 0.560       & 0.993       & 0.059          \\
        & FADG               & 90.1          & 100            & 91.7           & 98.4   & 2.537       & 0.541       & 0.995       & 0.105          \\
        \hline
\end{tabular}
\end{center}
\end{table*}

\section{EXPERIMENTS}
\label{sec:experiments}
The main objective with our experiments is to highlight how incorporating discriminator guidance improves the generative performance of the ARDM. Therefore, we have not put any effort into improving any architectural aspects of the backbone neural network, but use the same graph transformer as DiGress \citep{vignac_digress_2023} for both the generative model and the discriminator. Details on the architecture as well as details about training are given in the \suppmat.

As our model is working on partially masked graphs, we cannot compute the extra features used by \citet{vignac_digress_2023}, as we will not have access to, e.g., the graph Laplacian. On the other hand, we evaluate incorporating a recent approach by \citet{ekstrom_kelvinius_autoregressive_2023} where the order in the ARDM is not drawn from a uniform distribution. Instead, the graphs are generated either by first assigning the values of the nodes in a random order, and then the edges (called NsEs), or by always assigning the edges connecting the most recently assigned node with the already assigned nodes (called NEsN).

The evaluation metrics are computed on a sample of \thsnd{1} molecules generated by the models. In the SMC algorithms, we use $N=10$ particles per sample, meaning we effectively have to generate \thsnd{10} molecules, but only keep \thsnd{1} for evaluation. We include a model ARDM*, which is standard ARDM with a training time comparable with

\subsection{QM9}
We first evaluate our methods on the QM9 dataset \citep{wu_moleculenet_2018} and compare our results to DiGress \citep{vignac_digress_2023}, which is a recent diffusion model for graphs and hence a very competitive baseline. As the metrics on the standard QM9 setup (no hydrogens) are already very good with a standard ARDM (see \suppmat), we turn our attention to the more difficult task of explicitly modeling hydrogens. We evaluate our model using the metrics Validity (Val., fraction on molecules for which RDKit can obtain a valid SMILES string), Uniqueness (Uniq., fraction of molecules with a different SMILES representation), Atom stable (Atm.S, fraction of atoms with correct valency) and Molecule stable (Mol.S, fraction of molecules with 100 \% atom stable). As can be seen in \Cref{tab:qm9-results}, standard ARDM with a uniform order performs slightly worse on all metrics when compared to DiGress, but using a non-uniform order (NEsN or NsEs) improves performance. Most importantly, however, adding discriminator guidance improves over ARDM for all choices of orders, and in particular the SMC algorithms obtain close to ideal metrics when compared to the dataset.

\subsection{MOSES}
Next, we evaluate ARDM with discriminator guidance on MOSES \citep{polykovskiy_molecular_2020}, a more challenging dataset of almost two million small drug-like molecules. This dataset is a lot larger than QM9, and we therefore do not generate a new dataset of the same size as the original dataset, but instead only \thsnd{200} molecules, and then use random subsampling of these during training of the discriminator. As we do not see any direct differences in performance when using a different generation order, we present the results for the uniform order in \Cref{tab:moses-results}, and refer to the \suppmat for results when using different generation orders. 

MOSES is a standard benchmark with its own metrics, and apart from Validity and Uniqueness used for QM9, it also uses Novelty (fraction of the valid molecules not part of the training set), Filters (fraction of generated molecules which pass filters used in dataset construction), Fréchet ChemNet Distance (FCD) \citep{preuer_frechet_2018} (distance between hidden activations of a pretrained ChemNet for generated and test set), SNN (average similarity between molecules in generated set and their nearest neighbor in test set), Fragment Similarity (Frag, comparison of distributions of BRICS fragments \citep{degen_art_2008} in generated and test set), and Scaffold similarity (Scaff, comparison of frequencies of Bemis-Murcko scaffolds \citep{bemis_properties_1996} in generated and test set).

We observe that the different discriminator guidance methods improve many of the metrics over standard ARDM. However, it seems that they do so at a slight cost of novelty, i.e., they are slightly more prone to generating molecules from the training set. This could be interpreted as if the metrics are improved merely as a result of sampling more from the training set. However, this is not the case, since rerunning the evaluation using only novel molecules (see \suppmat), we observe the same improvements. It should also be noted that the metrics are computed over valid molecules, so even if the novelty is lower, the total number of valid \emph{and} novel molecules can be higher.

\section{DISCUSSION \& CONCLUSIONS}
In this paper, we have derived three discriminator guidance methods for ARDMs. 
First, we have a version (ARDG) that can be seen as a discrete counterpart of the continuous discriminator guidance by \citet{kim_refining_2023-1}. Second, we have two SMC versions (BSDG and FADG) that can further improve the generation quality by generating multiple interacting samples in parallel. Our empirical results when generating molecular graphs show how discriminator guidance improves the generative performance over standard ARDMs.

This improved performance naturally comes with an increased computational cost as the discriminator needs to be evaluated at each generation step. Additionally, for SMC we sample multiple particles in parallel for each generated sample. However, in contrast to other diffusion models, ARDMs have a fixed number of generation steps ($D$, one per variable). Choosing the number of particles $N$ can therefore be seen as a tuning knob that allows trading generation quality for computational cost, similarly to the number of diffusion steps in standard diffusion models (both continuous and discrete models like D3PM and DiGress). In particular, for the extreme case $N=1$, BSDG and FADG reduce to ARDM (no guidance) and ARDG, respectively, meaning that even for small $N$ we obtain well-functioning methods (which is confirmed by our empirical results with $N=10$).

We think that this trade-off can be of importance in applications or situations where it is more important that the obtained samples are of high quality, than being able to generate a large number of diverse samples with varying quality. For example, if the generated samples are to be further screened for some downstream application and this screening is expensive to perform, it is desirable that the samples that are chosen for screening are of higher quality, even if they are relatively few. SMC provides the possibility to optimize for this requirement by increasing the number of particles and lowering the number of generated samples within a fixed computational budget.

\acknowledgments{
This research is financially supported by the Swedish Research Council via the project
\emph{Handling Uncertainty in Machine Learning Systems} (contract number: 2020-04122),
the Wallenberg AI, Autonomous Systems and Software Program (WASP) funded by the Knut and Alice Wallenberg Foundation,
and
the Excellence Center at Linköping--Lund in Information Technology (ELLIIT).
The computations were enabled by the
Berzelius resource provided by the Knut and Alice Wallenberg
Foundation at the National Supercomputer Centre. 
}
\bibliography{references_discrete_diffusion}

\section*{Checklist}

\begin{enumerate}

 \item For all models and algorithms presented, check if you include:
 \begin{enumerate}
   \item A clear description of the mathematical setting, assumptions, algorithm, and/or model. \\\textbf{Yes}
   \item An analysis of the properties and complexity (time, space, sample size) of any algorithm. 
   \\ \textbf{Yes High-level description in main paper. Details in \suppmat.}
   \item (Optional) Anonymized source code, with specification of all dependencies, including external libraries.
   \\
   \textbf{After acceptance, code has been published on GitHub (see link on page 1)}
 \end{enumerate}

 \item For any theoretical claim, check if you include:
 \begin{enumerate}
   \item Statements of the full set of assumptions of all theoretical results. 
   \\
   \textbf{Yes}
   \item Complete proofs of all theoretical results.
   \\
   \textbf{Not applicable. We do not provide any theorem statements in need of proofs. All theoretical results are derived in the main text.}
   \item Clear explanations of any assumptions.\\
   \textbf{Yes. Assumptions explained in main paper, with additional details in \suppmat.}
 \end{enumerate}

 \item For all figures and tables that present empirical results, check if you include:
 \begin{enumerate}
   \item The code, data, and instructions needed to reproduce the main experimental results (either in the supplemental material or as a URL). \\
   \textbf{Yes}
   \item All the training details (e.g., data splits, hyperparameters, how they were chosen). \\
   \textbf{Yes, in \suppmat and code repository.}
    \item A clear definition of the specific measure or statistics and error bars (e.g., with respect to the random seed after running experiments multiple times). \\
    \textbf{Yes, we provide additional results to quantify the variability of the obtained results in the appendix}
    \item A description of the computing infrastructure used. (e.g., type of GPUs, internal cluster, or cloud provider). \\
    \textbf{Yes, in \suppmat}
 \end{enumerate}

 \item If you are using existing assets (e.g., code, data, models) or curating/releasing new assets, check if you include:
 \begin{enumerate}
   \item Citations of the creator If your work uses existing assets. 
   \\
   \textbf{Yes}
   \item The license information of the assets, if applicable. 
   \\
   \textbf{Yes, when applicable in \suppmat}
   \item New assets either in the supplemental material or as a URL, if applicable. 
   \\
   \textbf{Not Applicable}
   \item Information about consent from data providers/curators. 
   \\
   \textbf{Yes, when applicable in \suppmat}
   \item Discussion of sensible content if applicable, e.g., personally identifiable information or offensive content. 
   \\
   \textbf{Not Applicable}

 \end{enumerate}

 \item If you used crowdsourcing or conducted research with human subjects, check if you include:
 \begin{enumerate}
   \item The full text of instructions given to participants and screenshots. 
    \\
   \textbf{Not Applicable}
   \item Descriptions of potential participant risks, with links to Institutional Review Board (IRB) approvals if applicable. 
      \\
   \textbf{Not Applicable}
   \item The estimated hourly wage paid to participants and the total amount spent on participant compensation.
      \\
   \textbf{Not Applicable}
 \end{enumerate}

 \end{enumerate}

\appendix

\onecolumn

\section{SMC DETAILS}
\label{app:smc_details}
In this section, we give a more formal derivation of the two SMC algorithms that we use for discriminator guidance.

As mentioned in the main text, SMC samples from a sequence of target distributions, $\{\pi_t(\x_{\sigma(\leq t)})\}_{t=1}^D$, where $\pi_t$ is known up to a normalizing constans, i.e., 
\begin{align}
    \pi_t(\xonet) = \frac{\gammat}{Z_t}.
\end{align}

To obtain these samples, SMC consists of three steps: resampling, propagating, and weighting. Each of the three steps are associated with its own additional SMC ingredient: the \emph{ancestor probabilities} $\nu_{t-1}$ (used in the resampling step), the \emph{proposal distributions} $q_t(\xt | \xtone)$ (used in the propagation step), and the \emph{importance weights} $w_{t}$ (used in the weighting step). The latter can be obtained by first computing the unnormalized importance weights $\tilde{w}_t$ as 
\begin{align}
    \tilde{w}_t = \omega_t(\xonet) \frac{w_{t-1}}{\nu_{t-1}}, \label{eq:w_tilde_def}
\end{align}
where
\begin{align}
    \omega_t(\xonet) = \frac{\gammat}{\gammatone q(\xt | \xtone)},
\end{align}
and then normalizing by summing over all particles.

To design our SMC algorithms, we hence need to specify the target distributions $\gamma_t$, determine a proposal distribution, $q_t(\xt | \xtone)$, and how to compute the ancestor probabilities $\nu_{t-1}$. From this, the importance weights will follow according to \Cref{eq:w_tilde_def}.

We use $\gammat = W_t(\xonet) \ptheta(\xonet)$, as this leads to, at iteration $t=D$, samples that are approximately from $\pdata(\x)$ (see \Cref{eq:wt}).

\subsection{Bootstrap Particle Filter (BSPF)} For designing a bootstrap particle filter for discriminator guidance, we can use our pretrained model as the proposal distribution, i.e.,
\begin{align}
    q_t(\xt | \xtone) = \ptheta (\xt | \xtone).
\end{align}
As in a standard BSPF, we set the ancestor sample weights equal to the importance weights, i.e., 
\begin{align}
    \nu_{t-1} = w_{t-1}.
\end{align}
With this choice of proposal and ancestor probabilities, we have that
\begin{align}
    \omega_t(\xonet) &= \frac{\gammat}{\gammatone q(\xt | \xtone)} \nonumber \\
    &= \frac{W_t(\xonet) \ptheta(\xonet)}{W_{t-1}(\xtone) \ptheta(\xtone) \ptheta (\xt | \xtone)} \nonumber \\
    &= \frac{W_t(\xonet) \ptheta(\xonet)}{W_{t-1}(\xtone) \ptheta(\xonet)} \nonumber \\
    &= \frac{W_t(\xonet)}{W_{t-1}(\xtone)}.
\end{align}
Then, the importance weights become
\begin{align}
    \tilde{w}_t = \frac{W_t(\x_{\sigma(\leq t)})}{W_{t-1}(\x_{\sigma(<t)})} \frac{w_{t-1}}{\nu_{t-1}}, \label{eq:w_tilde_bspf}
\end{align}
which, if resampling is performed with $\nu_{t-1} = w_{t-1}$, gives $\tilde{w}_t = \frac{W_t(\xonet)}{W_{t-1}(\xtone)}$. However, if resampling is not performed (i.e., if $\ESS$ is sufficiently large), this can be viewed as instead setting $\nu_{t-1} = 1/N$ and using a low variance sampler (for example, systematic). The expression in \Cref{eq:w_tilde_bspf} covers both cases.

\subsection{Fully Adapted Particle Filter (FAPF)} 
In a fully adapted particle filter \citep{pitt_filtering_1999, NaessethLS:2019a}, the proposal should be the \emph{locally optimal proposal}

\begin{align}
    q_t(x_{\sigma(t)} | \x_{\sigma(<t)}) = \frac{\gamma_t (\{\xtone, \xt\})}{\int \gamma_t(\{\xtone, \xt\}) d\xt}.
\end{align}

For notational convenience when deriving the locally optimal proposal, we define
\begin{align}
    \Ct = \frac{\int \gamma_t(\{\xtone, \xt\}) d\xt}{\gammatone}.
\end{align}
By multiplying the right-hand side of the locally optimal proposal by $\frac{\gammatone}{\gammatone}$ and rearranging terms, we find that the locally optimal proposal is the same as the proposal in ARDG (\Cref{subsec:discr}), namely
\begin{align}
        q_t(\xt | \xtone) 
        &= \frac{\gammatone}{\int \gamma_t(\{\xtone, \xt \}) d\xt}\frac{\gamma_t (\{\xtone, \xt\})}{\gammatone} \nonumber \\
        &= \frac{1}{\Ct}\frac{\Wt \ptheta(\{\xtone, \xt\})}{\Wtone \ptheta(\xtone)} \nonumber \\
        &= \frac{1}{\Ct}\frac{\Wt}{\Wtone}\ptheta(\xt | \xtone).
    \end{align}
As $\Ct$ is part of the normalizing constant of the proposal, and as $x_{\sigma(t)}$ is discrete, we can compute $\Ct$ as
\begin{align}
    \Ct = \sum_{\xt} \frac{\Wt}{\Wtone}\ptheta(\xt | \xtone).  \label{eq:c_t_app}
\end{align}
In addition to using the locally optimal proposal, a fully adapted particle filter uses un-normalized ancestor probabilities which take into account possible values of future $\xt$ as \citep{pitt_filtering_1999}

\begin{align}
    \tilde{\nu}_{t-1} = w_{t-1} \frac{\int \gamma_t(\{\xtone, \xt\})d\xt}{\gammatone} = w_{t-1}\Ct,  \label{eq:nu_tilde}
\end{align}
which can be normalized by summing over all particles. 

Finally, the un-normalized importance weights are computed by using $\omega_t(\xonet)$, which with the locally optimal proposal becomes
\begin{align}
    \omega_t(\xonet) &= \frac{\gammat}{\gammatone q(\xt | \xtone)} \nonumber \\
    &= \frac{\gammat}{\gammatone}\frac{\Ct \gammatone}{\gammat} \nonumber \\
    &= \Ct,
\end{align}
resulting in
\begin{align}
    \tilde{w}_t = \frac{w_{t-1}}{\nu_{t-1}} \Ct. \label{eq:w_tilde_fapf}
\end{align}
It can be noted that if resampling is performed, this is done using $\nu_t$ obtained by normalizing $\tilde{\nu}_t$ from \Cref{eq:nu_tilde}, leading to $\tilde{w}_t = 1$ and hence, $w_t = 1 / N$. But just as for a BSPF, if $\ESS$ is still high enough and resampling is not performed in a step, this can be viewed as setting $\nu_t= 1/N$, meaning $\nu_t$ is not a normalization of $\tilde{\nu}_t$. We cover both these cases in \Cref{eq:w_tilde_fapf}.

\subsection{Experimental settings}
In all SMC experiments, we used $N=10$. We used systematic resampling and resampled when $\ESS < 0.7 N$. 

\section{NETWORK AND TRAINING DETAILS}
For both the generator and discriminator, we have used the graph transformer developed for DiGress. We have used the code from the the public DiGress code repository\footnote{\url{https://github.com/cvignac/DiGress/}}. However, we do not use the extra features used for DiGress, as we do not have access to a full graph. 

\subsection{Generator}
For the generator, we tried to closely follow the hyperparameters (e.g., number of layers and hidden dimensions) chosen for DiGress. When training, we used the Adam optimizer (while DiGress use AdamW), and tweaked the learning rate slightly depending on the dataset and generation order (Uniform, NEsN, or NEN). For QM9, we used 75 \% of the full dataset as training data (roughly \thsnd{98} molecules), slightly less than DiGress which used \thsnd{100} molecules. 

\subsection{Discriminator}
\paragraph{Architecture}
The discriminator has a backbone architecture that is identical to the generator (i.e., a graph transformer with the same number of layers and hidden dimensions). This backbone produces hidden features for each node and edge in the graph. In the generator, these features are used to predict the class logits in the generative distribution $p_{\theta}(x_{\sigma(t)} | \x_{\sigma(<t)})$. For the discriminator, however, we use mean pooling to obtain a pooled node feature, $\mathbf{h}_{\text{node}}$, and edge feature, $\mathbf{h}_{\text{edge}}$. That is, if letting $V$ and $E$ represent the set of nodes and set of edges, respectively, we obtain 
\begin{align}
    \mathbf{h}_{\text{node}} &= \frac{1}{|V|} \sum_{i \in V} \mathbf{h}_i \\
    \mathbf{h_{\text{edge}}} &= \frac{1}{|E|} \sum_{(i,j)\in E} \mathbf{h}_{(i,j)}.
\end{align}
These are concatenated and processed by an MLP to predict a single logit $l$, i.e.,
\begin{align}
    l = \MLP(\mathbf{h}_{\text{node}} \oplus \mathbf{h_{\text{edge}}}),
\end{align}
with $\oplus$ denoting concatenation of vectors. We used a single hidden layer in the MLP, with the hidden dimension being the same as the dimension of $\mathbf{h}_{\text{node}}$.

The logit can be converted into the probability of the sample being from the real dataset by applying a sigmoid. However, as shown in \Cref{eq:W_logit_space}, when computing the weights $W_t(\xonet)$, we use the logit directly as $W_t(\xonet) = \exp(l)$ which works better numerically. 

\paragraph{Training}
The already trained generative model is used to generate another dataset which is combined with the real dataset to train the discriminator. This generated dataset is of the same size as the original dataset in the case of QM9, and in the case of MOSES consists of \thsnd{200} samples, which is $\sim 10\%$ the size of the original dataset.

We used the AdamW optimizer and tweaked the learning rate by observing cross entropy on a validation set which consisted of both real and generated data. The validation set was created by setting aside $15 \%$ of the dataset. We find empirically that initializing the discriminator with the weights of the generator helps with training, especially in the initial phase. We therefore used this strategy when initializing all discriminators.

\subsection{Computational Resources}
Both generators and discriminators were trained on single NVIDIA A100 40 GB GPUs.

\section{ADDITIONAL RESULTS}
\subsection{QM9 without hydrogen}
\label{app:qm9_no_h_results}
For the task of generating molecules from QM9 when not modelling hydrogens, we provide the Validity and Uniqueness for standard ARDM, compared to DiGress, in \Cref{tab:qm9-no_h-results}.
\begin{table}[h]
\caption{Evaluation metrics on the QM9 dataset, without modeling hydrogens. The metrics are very good, as we have comparable results with DiGress \citep{vignac_digress_2023}.}
\label{tab:qm9-no_h-results}
\begin{center}
\begin{tabular}{l  c c  }
 & Validity & Uniqueness   \\
Model &  (\%)$\uparrow$ & (\%)$\uparrow$ \\
\hline\hline
DiGress & 99.0 & 96.2 \\
ARDM & 99.1 & 96.9  \\
\hline
\end{tabular}
\end{center}
\end{table}

\subsection{MOSES with non-Uniform Generation Order}
\label{app:moses_results}
Just as for QM9, we evaluated the approach of \citet{ekstrom_kelvinius_autoregressive_2023} where the distribution of generation orders, $p(\sigma)$, is not uniform over all possible orders. However, in contrast to QM9 where this approach gave significant boost in performance, this was not the case for MOSES. We do see, however, the same qualitative results that discriminator guidance consistently provides improved performance, given a choice of $p(\sigma)$. The full MOSES results can be found in \Cref{tab:moses-results-full}
\begin{table*}[ht]
\caption{\label{tab:moses-results-full}Evaluation metrics on the MOSES dataset, when using different generation orders and types of Discriminator Guidance (DG).}
\begin{center}
\begin{tabular}{clcccccccc}
        &                    & Validity      & Uniqueness     & Novelty        & Filters & FCD         & SNN         & Frag        & Scaf           \\
Order   & DG                 &(\%)$\uparrow$ & (\%)$\uparrow$ & (\%)$\uparrow$ & (\%)$\uparrow$ & $\downarrow$ & $\uparrow$ & $\uparrow$ & $\uparrow$  \\ 
\hline\hline
\multirow{4}{*}{\rotatebox[origin=c]{90}{Uniform}} 
        & ARDM                  & 82.2          & 100            & 97.2           & 94.9   & 2.937       & 0.483       & 0.993       & 0.067          \\
        & ARDM*           & 82.6          & 100            & 97.0           & 95.6   & 3.153         & 0.494       & 0.991       & 0.050          \\
        & ARDG            & 80.5          & 100            & 95.0           & 95.3   & 2.705       & 0.502       & 0.994       & 0.127          \\
        & BSDG               & 85.9          & 100            & 92.4           & 98.4   & 2.609       & 0.560       & 0.993       & 0.059          \\
        & FADG               & 90.1          & 100            & 91.7           & 98.4   & 2.537       & 0.541       & 0.995       & 0.105          \\
        \hline
\multirow{4}{*}{\rotatebox[origin=c]{90}{NEsN}}    
        & ARDM                  & 81.6          & 100            & 98.3           & 88.5   & 3.452       & 0.463       & 0.989       & 0.072          \\
        & ARDM*           & 82.1          & 100            & 98.6           & 94.5   & 2.942       & 0.486       & 0.986      & 0.065           \\
        & ARDG            & 85.0          & 100            & 97.2           & 95.3   & 3.138       & 0.495       & 0.990       & 0.072          \\
        & BSDG               & 87.2          & 100            & 96.7           & 96.7   & 2.681       & 0.519       & 0.993       & 0.064          \\
        & FADG               & 88.9          & 100            & 93.3           & 98.7   & 2.440       & 0.533       & 0.995       & 0.081          \\
        \hline
\multirow{4}{*}{\rotatebox[origin=c]{90}{NsEs}}    
        & ARDM                  & 81.3          & 100            & 97.0           & 93.5   & 2.993       & 0.494       & 0.986       & 0.039          \\
        & ARDM*           & 76.9          & 100             & 97.4          & 92.1   & 3.159       & 0.485       & 0.990       & 0.060         \\
        & ARDG            & 84.2          & 100            & 95.4           & 94.7   & 2.776       & 0.509       & 0.992       & 0.037          \\
        & BSDG               & 89.0          & 100            & 92.4           & 98.4   & 2.778       & 0.532       & 0.994       & 0.057          \\
        & FADG               & 91.0          & 100            & 93.7           & 98.8   & 2.343       & 0.539       & 0.995       & 0.075          \\
        \hline
\end{tabular}
\end{center}
\end{table*}

\subsection{MOSES metrics on only novel molecules}
\begin{table*}[ht]
\caption{Metrics for MOSES, but computed only on novel molecules \label{tab:moses_novel}}
\begin{center}
\begin{tabular}{clcccccccc}
        &                    & Valid        & Uniqueness     & Novelty        & Filters & FCD         & SNN         & Frag        & Scaf           \\
\hline\hline
\multirow{4}{*}{\rotatebox[origin=c]{90}{Uniform}} 
        & ARDM              & 82.2           & 100            & 100           & 94.7   & 2.972       & 0.480       & 0.993       & 0.068          \\
        & ARDG              & 80.5           & 100            & 100           & 95.0   & 2.797       & 0.496       & 0.993       & 0.131          \\
        & BSDG              & 85.9           & 100            & 100           & 98.2   & 2.699       & 0.524       & 0.994       & 0.063          \\
        & FADG              & 90.1           & 100            & 100           & 98.3   & 2.607       & 0.533       & 0.995       & 0.114          \\
        \hline
\end{tabular}
\end{center}
\end{table*}

In \Cref{tab:moses_novel}, we present the corresponding metrics for MOSES but only on novel molcules, and see the same pattern with DG improving over standard ARDM. As mentioned in the paper, all metrics are computed on only valid molecules, meaning that even if the novelty is lower, the \emph{total number} of novel (and valid) molecules generated with desired properties can be higher. For example, according to the numbers in \Cref{tab:moses_novel} and \Cref{tab:moses-results}, the fraction of the generated molecules by ARDM that are \textcolor{red}{valid}, \textcolor{blue}{novel} and pass the \textcolor{cyan}{filters} is $\textcolor{red}{0.822}\cdot \textcolor{blue}{0.972} \cdot \textcolor{cyan}{0.947} = \textcolor{orange}{75.7} \%$, while the same number for FADG is $\textcolor{red}{0.901} \cdot \textcolor{blue}{0.917} \cdot \textcolor{cyan}{0.983} = \textcolor{orange}{81.2} \%$.

\subsection{Generation with different seeds}
To evaluate how much the results can vary, we generate new sets of \thsnd{1} molecules using a different seed, and present the results in \Cref{tab:qm9-results_seed} and \Cref{tab:moses-results-full_seed}. The same pattern can be seen in these tables as those in the main text, with discriminator guidance improving most metrics.

\begin{table*}[ht]
\caption{\label{tab:qm9-results_seed}Evaluation metrics on the QM9 dataset when using a different seed}
\begin{center}
\begin{tabular}{clcccc}
                        &            & Validity      & Uniqueness     & Atom stable    & Mol stable        \\
Order                   & Model      & (\%)$\uparrow$& (\%)$\uparrow$ & (\%)$\uparrow$ & (\%)$\uparrow$    \\ 
\hline\hline
                        & Test data    & 97.9          & 100            & 98.6           & 87.4              \\ 
                        & DiGress    & 95.4          & 97.6           & 98.1           & 79.8              \\
                        \hline
\multirow{4}{*}{\rotatebox[origin=c]{90}{Uniform}} 
    & ARDM              & 87.5     & 100       & 96.0       & 55.3       \\
    & ARDG              & 88.0     & 99.4       & 97.0       & 67.8       \\
    & BSDG              & 97.0     & 99.7       & 98.9       & 89.0       \\
    & FADG              & 97.6     & 99.2       & 98.7       & 88.3       \\
    \hline
\multirow{4}{*}{\rotatebox[origin=c]{90}{NEsN}}    
    & ARDM              & 95.1     & 99.7       & 97.1       & 73.0       \\
    & ARDG              & 93.2     & 99.8       & 97.3       & 74.5       \\
    & BSDG              & 98.8     & 99.7       & 98.8       & 89.4       \\
    & FADG              & 97.0     & 99.6       & 98.6       & 88.0       \\
    \hline
\multirow{4}{*}{\rotatebox[origin=c]{90}{NsEs}}    
    & ARDM              & 92.7     & 99.9       & 97.3       & 77.0       \\
    & ARDG              & 94.3     & 99.7       & 97.8       & 81.6       \\
    & BSDG              & 97.9     & 99.7       & 98.5       & 87.1       \\
    & FADG              & 97.3     & 99.1       & 98.4       & 85.8       \\
    \hline
\end{tabular}
\end{center}
\end{table*}

\begin{table*}[ht]
\caption{\label{tab:moses-results-full_seed}Evaluation metrics on the MOSES dataset with a different seed.}
\begin{center}
\begin{tabular}{clcccccccc}
        &                    & Validity      & Uniqueness     & Novelty        & Filters & FCD         & SNN         & Frag        & Scaf           \\
Order   & DG                 &(\%)$\uparrow$ & (\%)$\uparrow$ & (\%)$\uparrow$ & (\%)$\uparrow$ & $\downarrow$ & $\uparrow$ & $\uparrow$ & $\uparrow$  \\ 
\hline\hline
\multirow{4}{*}{\rotatebox[origin=c]{90}{Uniform}} 
        & ARDM               & 80.4          & 100            & 97.4           & 93.8   & 2.942       & 0.487       & 0.992       & 0.095          \\
        & ARDG               & 84.6          & 100            & 95.0           & 94.9   & 2.503       & 0.509       & 0.995       & 0.106          \\
        & BSDG               & 89.6          & 100            & 92.4           & 98.2   & 2.789       & 0.527       & 0.992       & 0.050          \\
        & FADG               & 88.3          & 100            & 91.4           & 98.2   & 2.532       & 0.537       & 0.993       & 0.082          \\
        \hline
\multirow{4}{*}{\rotatebox[origin=c]{90}{NEsN}}    
        & ARDM               & 81.9          & 100            & 98.2           & 90.2   & 3.110       & 0.468       & 0.990       & 0.083          \\
        & ARDG               & 85.6          & 100            & 97.1           & 94.2   & 2.887       & 0.501       & 0.993       & 0.075          \\
        & BSDG               & 86.6          & 100            & 95.4           & 97.2   & 2.746       & 0.518       & 0.994       & 0.046          \\
        & FADG               & 89.2          & 100            & 93.4           & 98.1   & 2.749       & 0.527       & 0.993       & 0.035          \\
        \hline
\multirow{4}{*}{\rotatebox[origin=c]{90}{NsEs}}    
        & ARDM               & 78.8          & 100            & 98.6           & 93.3   & 3.159       & 0.487       & 0.987       & 0.091          \\
        & ARDG               & 83.7          & 100            & 96.3           & 95.5   & 2.736       & 0.505       & 0.994       & 0.055          \\
        & BSDG               & 88.0          & 100            & 93.2           & 98.0   & 2.636       & 0.526       & 0.995       & 0.079          \\
        & FADG               & 91.0          & 100            & 92.7           & 98.7   & 2.565       & 0.535       & 0.993       & 0.095          \\
        \hline
\end{tabular}
\end{center}
\end{table*}

\subsection{Varying number of particles}
In \Cref{tab:n-results}, we present the results on MOSES for BSDG and FADG when varying the number of particles, $N$. We conclude that $N=10$ is a reasonable compute-quality trade-off for this case.

\begin{table*}[ht]
\caption{\label{tab:n-results}Evaluation metrics on the MOSES dataset when using the uniform generation order and varying the number of particles, $N$.}
\begin{center}
\begin{tabular}{clcccccccc}
   &                    & Validity      & Uniqueness     & Novelty        & Filters & FCD         & SNN         & Frag        & Scaf           \\
   & $N$                 &(\%)$\uparrow$ & (\%)$\uparrow$ & (\%)$\uparrow$ & (\%)$\uparrow$ & $\downarrow$ & $\uparrow$ & $\uparrow$ & $\uparrow$  \\ 
\hline\hline
\multirow{4}{*}{\rotatebox[origin=c]{90}{BSDG}} 
        &  5            & 87.9          & 100            & 92.0           & 97.5   & 2.606       & 0.528       & 0.993       & 0.073          \\
        &  10           & 85.9          & 100            & 92.4           & 98.4   & 2.609       & 0.560       & 0.993       & 0.059          \\
        &  25           & 87.9          & 100            & 92.4           & 98.2   & 2.650       & 0.538       & 0.992       & 0.079          \\
        &  50           & 89.8          & 100            & 92.9           & 98.4   & 2.459       & 0.537       & 0.993       & 0.090          \\
        \hline
\multirow{4}{*}{\rotatebox[origin=c]{90}{FADG}}
        & 5             & 88.1          & 100            & 91.5           & 98.4   & 2.504       & 0.533       & 0.994       & 0.082          \\
        & 10            & 90.1          & 100            & 91.7           & 98.4   & 2.537       & 0.541       & 0.995       & 0.105          \\
        & 25            & 90.1          & 100            & 92.1           & 99.1   & 2.589       & 0.537       & 0.991       & 0.036          \\
        & 50            & 87.5          & 100            & 90.5           & 99.1   & 2.809       & 0.532       & 0.993       & 0.035          \\
        \hline
\end{tabular}
\end{center}
\end{table*}

\section{LICENSE INFORMATION}
QM9 is provided by the original authors on Figshare\footnote{\url{https://springernature.figshare.com/collections/Quantum_chemistry_structures_and_properties_of_134_kilo_molecules/978904}}, and was accessed through \texttt{Pytorch Geometric}\footnote{\url{https://pytorch-geometric.readthedocs.io/en/latest/generated/torch_geometric.datasets.QM9.html}}.
The license is unspecified.

The MOSES dataset was accessed via the official open source library \texttt{moses} \footnote{\url{https://github.com/molecularsets/moses/}} with an MIT License. 

Parts of the code used in this work (e.g., the graph transformer and code for evaluation) were taken from the public DiGress code repository\footnote{\url{https://github.com/cvignac/DiGress/}} and modified. This code is licensed with an MIT License.

\end{document}